\documentclass[letterpaper, 10 pt, conference]{ieeeconf}  

\IEEEoverridecommandlockouts                              

\usepackage{multicol}
\usepackage[bookmarks=true]{hyperref}

\usepackage{graphics} 
\usepackage{epsfig} 
\graphicspath{{figures_final/}}
\usepackage[algoruled, vlined]{algorithm2e}
\usepackage{times} 
\usepackage{amsmath} 
\usepackage{amssymb}  
\usepackage{multirow}
\usepackage{subfigure}
\usepackage{tabularx}
\usepackage{color}
\usepackage{xcolor}

\usepackage{etoolbox}

\providetoggle{robotresults}
\settoggle{robotresults}{true} 
\providetoggle{validation}
\settoggle{validation}{false}

\newcommand{\Jim}[1]{#1}

\title{\LARGE \bf
Goal Set Inverse Optimal Control and Iterative Re-planning for Predicting Human Reaching Motions in Shared Workspaces \vspace{-0.1in}
}

\author{Jim Mainprice$^{1}$, Rafi Hayne$^{2}$, Dmitry Berenson$^{2}$ \\
\textit{\small{$\{$jim.mainprice@tuebingen.mpg.de , rhhayne@wpi.edu, dberenson@cs.wpi.edu$\}$}}\\
\textit{\small{ $^{1}$ Max-Planck-Institute for Intelligent Systems, Autonomous Motion Department, Paul-Ehrlich-Str. 15, 72076 T\"{u}bingen, Germany }}\\
\textit{\small{ $^{2}$Robotics Engineering Program, Worcester Polytechnic Institute, 100 Institute Rd, Worcester, MA 01609. }}
\vspace{-0.5cm}
\thanks{
This work is supported in part by the Office of Naval Research under Grant N00014-13-1-0735 and by the National Science Foundation under Grant IIS-1317462.
}
}

\begin{document}

\maketitle
\thispagestyle{empty}
\pagestyle{empty}

\begin{abstract}

To enable safe and efficient human-robot collaboration in shared workspaces it is important for the robot to predict how a human will move when performing a task. While predicting human motion for tasks not known \textit{a priori} is very challenging, we argue that single-arm reaching motions for known tasks in collaborative settings (which are especially relevant for manufacturing) are indeed predictable. Two hypotheses underlie our approach for predicting such motions: First, that the trajectory the human performs is optimal with respect to an unknown cost function, and second, that human adaptation to their partner's motion can be captured well through iterative re-planning with the above cost function. 
The key to our approach is thus to learn a cost function which ``explains'' the motion of the human. To do this, we gather example trajectories from pairs of participants performing a collaborative assembly task using motion capture. 
We then use Inverse Optimal Control to learn a cost function from these trajectories. 
Finally, we predict reaching motions from the human's current configuration to a task-space goal region by iteratively re-planning a trajectory using the learned cost function. Our planning algorithm is based on the trajectory optimizer STOMP \cite{Kalakrishnan:11}, it plans for a 23 DoF human kinematic model and accounts for the presence of a moving collaborator and obstacles in the environment.
Our results suggest that in most cases, our method outperforms baseline methods when predicting motions. We also show that our method outperforms baselines for predicting human motion when a human and a robot share the workspace. 



\end{abstract}

\section{Introduction}

Human-robot collaboration is increasingly studied in an industrial context because many tasks (such as electronics or aircraft assembly) are stressful for humans but have proven difficult to automate. In such cases the human and the robot workers must adapt to each others' decisions and motions. In this paper we address an important step toward more fluid human-robot collaboration: the ability to predict human motion in a shared workspace.

A great deal of work in the fields of neuroscience \cite{Flash:85, Biess:07, Todorov:02} and biomechanics \cite{Wu:05} has sought to model the principles underlying human motion. However, human motion in environments with obstacles has been difficult to characterize. Furthermore, human motion in collaborative tasks where two humans share a workspace is difficult to model due to unclear social, interference, and comfort criteria. While some of these principles have been studied in the context of human \textit{navigation} \cite{Kruse:13}, to our knowledge very few works address the problem of predicting human collaborative \textit{manipulation} tasks \cite{Rozo:13}, and no framework exists for predicting such human motion among obstacles.

\begin{figure}[t]
      \center
      \includegraphics[width =0.45\linewidth]{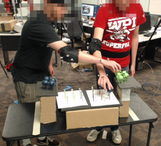}
      \includegraphics[width =0.51\linewidth]{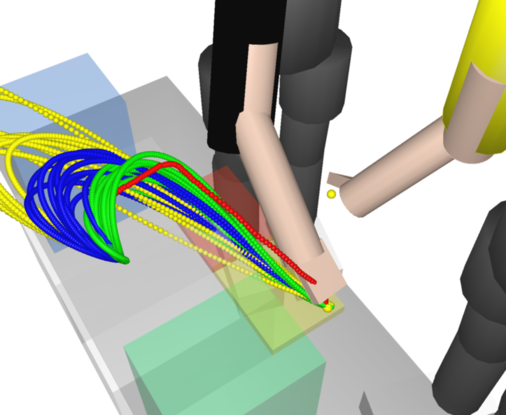}
      \caption{Shared workspace assembly experiment (left) and Trajectories predicted by Goal Set STOMP, using IOC in green, and manually tuned cost functions in yellow and blue (right). 
      }
      \label{fig:collaborative}
      \vspace{-.6cm}
\end{figure}

{\Jim
This paper presents such a framework for reaching motions, which is based on studying how two humans collaborate in a shared workspace (as in Figure \ref{fig:collaborative}). This is an important category of motions to be able to predict, since many pick-and-place tasks in manufacturing fall into this category. Being able to predict these motions well can move us closer to enabling safe and efficient human-robot collaboration.
}

%

Studying how two humans collaborate also gives us an important baseline against which human-robot collaborations can be judged; if we can predict what a natural motion for a human is in a given collaborative context, we can judge when the human deviates significantly from that motion in response to a robot's actions. We can also gauge how well a human is acclimated to a robot collaborator. This paper focuses on the method to obtain an accurate prediction for the above purpose, and though we envision eventually using this prediction in the robot's motion planner, this application is not within the scope of this paper.

Our approach is based on two hypotheses about collaborative human motion: 1) The trajectory the human performs is optimal with respect to an unknown cost function, and 2) Human adaptation to their partner's motion can be captured well through iterative re-planning of a trajectory which is locally-optimal with respect to the same cost function. Our method thus seeks to learn a cost function for which the human's motion is locally-optimal from training data.

To gather training data, we record the motion of two humans performing a collaborative task using a motion capture system and then manually-segment that recording into individual reaching motions. These reaching motions, along with a set of feature functions encoding trajectory smoothness and distance relationships between the humans are used as input for the Path Integral Inverse Reinforcement Learning (PIIRL) algorithm \cite{Kalakrishnan:13}. PIIRL produces a weighting for the feature functions that captures their relative importance. The learned cost function is then a weighted sum of the feature functions using the learned weights. To predict human motion we input the learned cost function into a trajectory optimization algorithm based on STOMP \cite{Kalakrishnan:11}. We make two changes to the algorithm, which are crucial for our domain: 1) We adapt it for iterative motion re-planning in a dynamic environment; and 2) We allow the algorithm to search over a task-space goal region instead of specifying a goal configuration. The second change is especially important in predicting human motion, as we do not know the goal configuration \textit{a priori}.

In our experiments we gathered the training data from pairs of participants in a structured assembly task (see Figure \ref{fig:collaborative}). We found that we are able to capture a cost function for collaborative reaching motions that outperforms baseline methods in most cases. We also found that re-planning was more effective than single-shot planning for capturing a human's adaptation to their partner's motion in cases where the motion of the two participants interfered significantly. Finally, we show that our method can be used to predict human motion when a human and a robot share the workspace better than baseline methods. 


The remainder of this paper is structured as follows: In the next section we give a description of related work. In Section \ref{sec:approach} we describe the approach that enables us to recover the cost function from training data. In Section \ref{sec:experiment}, we present the experimental setup used to gather collaborative reaching motions. Section \ref{sec:results} presents results that illustrate the ability of our method to predict collaborative reaching motions.
\iftoggle{robotresults}{We then present results from the human-robot workspace sharing scenario in Section \ref{sec:humanrobot}.}{}

A preliminary version of this work appeared in \cite{Mainprice:15}. The version presented here contains changes to the trajectory optimization and inverse optimal control algorithms that allow predicting the human's motion with a task-space goal set, which is essential for real-world applications where the human's goal configuration is unknown. We also present results from an expanded human subjects study, results on generalizing learned weight vectors to new goal regions and among participants, and the results of a human-robot workspace sharing experiment. 

\section{Related Work}
\label{sec:related_work}


\subsection{Probabilistic Graphical Models}

{\Jim
Graphical models have often been used for predicting human motion.
Motion prediction based on Gaussian Mixture Models (GMMs), commonly used in gesture recognition \cite{Wilson:99}, has been shown to perform well for high-dimensional movement recognition. 
Hidden Markov Models (HMMs), another popular stochastic modeling technique for human motion recognition and prediction \cite{Bretzner:02}, were used in \cite{Kulic:12}, where Kuli\c{c} et al. describe an approach for online incremental learning of full body motion primitives from observation of human motion, allowing the same model to be used for both motion recognition and motion generation. Finally, Conditional Random Fields (CRFs) were used in \cite{Koppula:13}, where Koppula and Saxena predict 3D trajectories of the human hand based on affordances. This work was recently extended in \cite{Jiang:14} to predict high-dimensional trajectories.
}

While these graphical model representations (i.e., GMMs, HMMs, and CRFs) allow one to efficiently encode relationships such as those between activities, objects and motions, they do not capture obstacles constraints well, an issue we address in this work. We also show that our method outperforms the GMM approach we have employed in \cite{Mainprice:13} for the collaborative reaching motions we are considering.

\subsection{Optimal Control}

{\Jim
Optimal control has been investigated for decades and recently Ganesh and Burdet \cite{Ganesh:13} used a manipulation task to show that the Central Nervous System (CNS)} uses a motion planning phase with multiple plans, and a memory mechanism.
Many experiments investigating reaching under various conditions \cite{Flash:85,Biess:07} suggest that at a high level the human motor-behavior can be modeled by the minimization of a cost function used to weigh different movement options for a task, as well as to select a particular solution.
Stochastic Optimal Control \cite{Bertsekas:95} provides a theoretical framework for these models while taking into account motor noise inherent to sensorimotor control \cite{Todorov:02}. In this spirit, a detailed subject-customized bio-mechanical model has been used in \cite{Demircan:10} to efficiently reconstruct a subject's motion dynamics from motion capture data in real-time using a whole-body control approach. 

These works suggest that an optimality criterion can model human motor behavior, the aim of our work is to find such a criterion to predict human reaching motions in shared workspaces, without resorting to musculoskeletal modeling of the human such as \cite{Demircan:10}.

\subsection{Inverse Optimal Control}
The Inverse Optimal Control (IOC) problem, occasionally named Inverse Reinforcement Learning (IRL), is the problem of finding the cost or reward function that an agent optimizes when computing a trajectory or policy given a set of demonstrated solutions. It is usually framed in the context of a Markov Decision Process. IRL was introduced by Ng et al. in \cite{Ng:00}, who proposed two algorithms for discrete and continuous states spaces. Later, apprenticeship learning \cite{Abbeel:04} introduced the idea of maximizing the margin between the cost of the demonstration and other solutions. Apprenticeship learning consists of iteratively solving the forward problem by modifying the weights at each iteration. 

In \cite{Ziebart:08}, Ziebart et al. proposed an approach to IRL based on the maximum entropy principle. {\Jim The recent methods based on this formulation \cite{Boularias:11, Aghasadeghi:11, Park:13b} do not require solving the forward problem and allow handling high-dimensional continuous state spaces. Instead of solving the forward problem they either sample trajectories or solve for the local optimality of the demonstrations using the demonstrations' feature-derivatives with respect to states and actions.} Sampling-based IOC approaches generally allow solving model-free problems. They are particularly efficient when using motion primitives such as in \cite{Boularias:11}. Motion primitives reduce the action space dimensionality, and allow learning of closed loop behavior \cite{Stulp:13}. Our approach derives from the sampling-based algorithm introduced in \cite{Kalakrishnan:13}. This method only requires local optimality of the demonstrated trajectories and can be used in a model-based fashion.

Only a few studies have employed IOC approaches to determine objective functions for the optimal control problem for human motion generation \cite{Albrecht:11,Berret:11,Sylla:14}. 

These objectives are usually of a kinematic, dynamic or geodesic nature.
There is a question as to whether IOC can answer fundamental scientific questions about human movement. Since IOC can only inform us about the comparative influence of the selected basis functions, if the true criterion is not modeled, the result is not informative. 
However, our goal is not to find the true cost function used by a human, rather it is to create a practical model of human reaching that can be used for prediction in collaborative tasks.

\subsection{Trajectory Optimization}
We rely on recent developments in trajectory optimization for motion planning \cite{Zucker:13, Kalakrishnan:11} to compute low-cost motion predictions. Our trajectory optimizer is based on the Stochastic Trajectory Optimizer for Motion Planning (STOMP) algorithm, which has proven effective for the type of manipulation motion planning we consider \cite{Kalakrishnan:11}. Recently, STOMP was adapted to run faster than real-time \cite{Park:13}. We plan to employ this new method in future work.


\begin{figure*}[t]
      \center
      \includegraphics[width=1.0\linewidth]{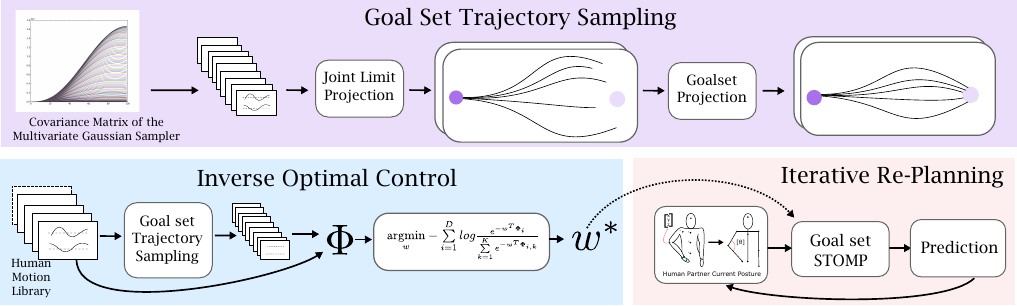}
      \vspace{-.5cm}
      \caption{Data flow through the system. The gathered human motion library is used to generate sample trajectories. Features ($\Phi$) are then computed for the demonstrated and sampled trajectories. The PIIRL algorithm is then applied to generate a weight vector $w^*$. Prediction of collaborative human reaching motions can then be performed by an iterative re-planning algorithm based on STOMP, relying on the learned weight vector $w^*$ and a kinematic model of the human. We use goal set trajectory optimization in both IOC and Iterative Re-Planning.}
      \label{fig:pipeline}
      \vspace{-.6cm}
\end{figure*}

\section{Approach}
\label{sec:approach}

Our approach to predicting human motion in collaborative manipulation tasks consists of two phases (see Figure \ref{fig:pipeline}). First we gather a library of collaborative motions. We then segment the motions into elementary reaching motions (i.e., from a resting configuration to a grasping configuration). The obtained trajectory library is used as demonstrations for the IOC algorithm to learn a cost function. Finally, we use the learned cost function inside an iterative motion re-planner to predict how the human will move in close proximity to another human or robot.

\subsection{Inverse Optimal Control algorithm}
\label{sec:IOC}

Intuitively, solving IOC consists of finding a cost function under which the demonstrated behavior is optimal. In most of the state-of-the-art techniques \cite{Ng:00,Abbeel:04,Ziebart:08,Kalakrishnan:11}, the cost function for a given trajectory $\xi$ has been parameterized by a linear combination of user defined features  $C(\xi) = w^T \Phi(\xi)$, where $w$ is the parameter of interest and $\Phi$ the multi-valued feature function.

A human reaching motion trajectory can be represented as a time-parameterized curve in some human configuration space. These curves can be discretized in sequences of waypoints (i.e., configurations) at evenly spaced time intervals, leading to the following definition for a trajectory:
$$
\xi = 
\left[
\begin{array}{c}
q_1 \;
\hdots \;
q_N
\end{array}
\right]^T
$$

\noindent
where $q_i$ are row vectors of configurations. The dimensionality of the corresponding vector space is $M * N$, where $M$ is the number of DoF of the kinematic model, and $N$ the number of waypoints.

Human reaching motions are inherently high-dimensional (in this work we consider $M =$ 23 DoF and $N =$ 100 waypoints for a duration of approximately one second). 
Computing globally-optimal solutions for motion planning problems of this nature is known to be intractable as the obstacles geometry generally introduces multiple local minima and the problem can not assumed to be convex, thus guaranteeing a globally-optimal IOC solution is also intractable. Furthermore, when we observe human motion using a motion capture system we do not have access to the dynamics of the human motor behavior. Hence in this work we focus on learning kinematic motion policies, i.e., we do not model forces or torques and we only parameterize the cost function with kinematic quantities. This simplification allows us to use a model-based approach (i.e., we assume that the transitions between states are all known and deterministic).

To solve the IOC problem locally we use the sampling-based \textit{Path Integral Inverse Reinforcement Learning} (PIIRL) algorithm \cite{Kalakrishnan:13}, which can deal with high-dimensional continuous state-action spaces, and only requires local optimality of the demonstrated trajectories. \footnote{PIIRL can also be used for model-free IOC, which typically requires rollouts on the physical system and is thus inapplicable when learning from motion capture data.}
Compared to other IOC methods, PIIRL uses a prior on the kinematic property of the motion behavior to be learned by sampling from a fixed distribution. This allows the algorithm to be less myopic when generating cost functions. Other sampling-based IOC techniques typically use importance sampling \cite{Boularias:11}. PIIRL has been shown to outperform other state-of-the-art methods in \cite{Kalakrishnan:13} for kinesthetically taught manipulation motions, which are similar to the motions we consider in this work.

The problem considered by PIIRL is to recover a cost function composed of a control cost, and a general cost (i.e., configuration dependent) term that can be combined with a terminal cost, which we do not use in this work. Instead we use goal set trajectory sampling as described in Section \ref{sec:goal_set}. 

The cumulative cost $C(\xi)$, is a linear combination of user defined features $
\Phi(\xi) = 
    \begin{bmatrix}
       G(\xi), \;
       A(\xi) 
     \end{bmatrix}^T
$,
where $A$ is the term enforcing smoothness (i.e., control cost) and $G$ the general term of the form:
$$ G(\xi) = \int_{t=0}^T \! \phi(q_t) \, \mathrm{d}t \simeq \sum \limits_{i=1}^N \phi(q_i) \Delta t ,$$
\noindent
where $q_i$ is the configuration at index $i$ along the trajectory and $N$ the number of waypoints. Defining the cost function as a linear combination of features makes the problem of learning the weight vector $w$ tractable.

Thus each feature function penalizes motions that do not respect an associated property, see Section \ref{sec:features} for a description of the features we use to predict human motion.

PIIRL samples trajectories with low smoothness features around each demonstration in order to estimate the partition function \cite{Ziebart:08}. The sampling distribution is defined using Multivariate Gaussians $\mathcal{N}(\xi_d, \Sigma = \sigma \mathbf{R}^{-1})$, centered at each demonstration $\xi_d$, where $\mathbf{R} = K^T K$, and $K$ is a matrix of finite differences that computes time derivatives of configurations along the trajectory ($K \xi_d$).
{\Jim
We set $K$ to sample trajectories with low sums of accelerations, which, for the one dimensional case, has a band diagonal structure of the following form:
$$
 K = 
 \left[
 \begin{smallmatrix}
       6  & -4	& 1  & \dots & 0 & 0 & 0 \\
       -4 & 6 & -4  & \dots & 0 & 0 & 0 \\
        1 & -4 & 6  & \dots & 0 & 0 & 0 \\
         & \vdots &  & \ddots &  & \vdots &  \\
        0 & 0 & 0  & \dots & 6 & -4 & 1 \\
        0 & 0 & 0  & \dots & -4 & 6 & -4 \\
        0 & 0 & 0  & \dots & 1 & -4 & 1 \\
  \end{smallmatrix}
  \right].
$$
\noindent
To learn cost functions that allow planning towards a task-space goal set, i.e., where the end configuration $q_N$ is not specified, we introduced a modified distribution discussed in Section \ref{sec:goal_set} and call the resulting algorithm Goalset-PIIRL.
}
The weights are then obtained by solving the following convex minimization problem:
\begin{equation}
{\color{black} 
w^* = \underset{w}{\operatorname{argmin}}  - \sum \limits_{i=1}^D log \frac{e^{-w^T \Phi_i}}{ \sum \limits_{s=1}^S e^{-w^T \Phi_{i,s} }} ,
}
\label{eq:loss}
\end{equation}

\noindent
where $D$ is the number of demonstrations and $S$ the number of trajectory samples per demonstration. $\Phi_i$ are the features computed for demonstration $i$ and $\Phi_{i,s}$ for the trajectory samples around that demonstration.

Sampling from $\mathcal{N}(\xi_d, \Sigma = \sigma \mathbf{R}^{-1})$ allows the algorithm to converge with less samples than other methods (see \cite{Kalakrishnan:13}).
Note that in the sampling phase we change the last entry of the block of $K$ corresponding to each DoF to allow variation of the end configuration $q_N$ in the trajectory samples as shown in the upper part of Figure \ref{fig:pipeline}. We then perform joint limit and goal set projection with respect to the metric $\mathbf{R}$ (see Section \ref{sec:goal_set}). Trajectory samples colliding with the environment and the other human are discarded by performing collision detection.

In the original version of PIIRL, a penalty on the $L_1$ norm of the weight vector $w$ is added to the loss function in equation \ref{eq:loss} to achieve learning with a large set of features. In this case the loss function is still convex but non differentiable due to the regularization term. In order to handle this non linearity, the Orthant-Wise Limited-memory Quasi-Newton \cite{Andrew:07} algorithm is used, which introduces additional projection steps and constrains the search to one orthant at a time. Using a regularization term adds a supplementary parameter to the algorithm that can be tuned through cross validation. In order to tune the regularizer we run a learning phase with a range of values and select the one that minimizes our validation criterion.

\subsection{Iterative re-planning}
\label{sec:dyanmic}

\begin{figure}[t]
      \center
      \includegraphics[width =0.8\linewidth]{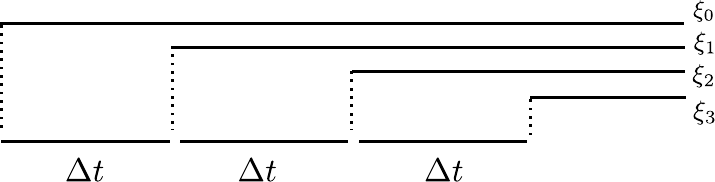}
      \caption{Division of a demonstration $\xi_0$ into smaller segments}
      \label{fig:division}
      \vspace{-0.7cm}
\end{figure}

Iterative re-planning consists of planning iteratively while considering the current environment as static. It is a common approach to account for dynamic obstacles in robot motion planning \cite{Brock:02,Park:13}. Typical approaches either maintain a tree or graph of collision-free motions, which is updated at each replanning step, or deform the current trajectory locally given the updated positions of obstacles in the world. Our approach aims to recover a cost function that can be used for such a framework. Thus, once the library of collaborative motion trajectories is gathered, it is segmented manually in elementary manipulation motions, which are then cut in smaller segments by advancing $\Delta t$ along each demonstration $\xi_0$ as depicted in Figure \ref{fig:division}. The newly generated sub-segments are added to the demonstration-trajectory set.
For each segment the initial velocity $\dot{q_0}$, acceleration $\overset{..}{q_0}$ and jerk $\overset{...}{q_0}$, as well as the configuration of the other human and the positions of obstacles are used to compute the features for that segment and for its corresponding sample trajectories.

\subsection{Goal Set STOMP}
\label{sec:STOMP}

When planning with the human model, we use the STOMP algorithm \cite{Kalakrishnan:11}, which is a trajectory optimizer that iteratively deforms an initial solution where the initial and goal configuration are fixed, by stochastically estimating the gradient in trajectory space.
At each iteration, trajectories are sampled from a Multivariate Gaussian distribution $\mathcal{N}(\xi, \Sigma = \sigma \mathbf{R}^{-1})$, and combined to generate the update. Thus, STOMP does not require the analytical gradient of the cost function to be known, and generally converges to a local minimum within 100 iterations. Our goal set version of the algorithm, which is similar in spirit to \cite{Dragan:11}, relies on sampling trajectories with different goal configurations conditioned to meet the goal region (see Section \ref{sec:goal_set}). In order to initialize the optimizer we use Jacobian-based inverse kinematics \cite{Siciliano:10} to seed the algorithm with an initial goal configuration (analytical methods cannot be applied due to the redundancy of the human kinematics). This method finds an inverse kinematics solution that minimizes the configuration-space euclidean distance to the initial configuration.

The original STOMP algorithm presented in \cite{Kalakrishnan:11} optimizes a combination of obstacle and smoothness cost. The first is estimated by summing a penetration cost for a set of bounding spheres to the obstacles at every waypoint using a signed Euclidean Distance Transform (EDT) as defined in \cite{Zucker:13}; the second is estimated by summing the squared accelerations along the trajectory using finite differencing (Equation \ref{eq:acc}).
In our version of the algorithm, we use a richer set of smoothness features, as we also account for task-space smoothness, described in \ref{sec:features}.
In order to account for smoothness at each re-planning step, a buffer of configuration waypoints from the previous re-planning step is used to compute velocity, acceleration, and jerk at the initial configuration. 
Finally, to account for the posture and for the other human, we add a third cost criterion defined in section \ref{sec:features}. Note that the weight of the obstacle cost is manually tuned in our result section. 

\subsection{Goal Set Trajectory Sampling}
\label{sec:goal_set}

Here we introduce goal set trajectory sampling used in our version of \textit{Goalset-STOMP} and \textit{Goalset-PIIRL}. In the standard version of both algorithms, trajectories are sampled from a Multivariate Gaussian distribution $\mathcal{N}(\xi, \Sigma = \sigma \mathbf{R}^{-1})$, that generates trajectories with fixed end configuration. $\xi$ is the trajectory we are considering, which is the current solution when planning, and demonstration when learning.

In order to sample trajectories that meet the goal set constraint while allowing different postures at the goal configuration, we sample from a modified covariance matrix (see Figure \ref{fig:pipeline} for its one dimensional version) and we project the samples to the goal set with respect to the metric $\mathbf{R}$. We define goal set constraints as having the final configuration of the motion $q_N$ place the human's hand at a given point. \footnote{We restrict our definition of goal sets to ones specified by a 3D point with free rotation, however note that a similar algorithm can be derived for a broader class of task space regions as defined in \cite{Berenson:11}.}

The modification to the covariance matrix can be obtained by changing the endpoint smoothness term computation in the finite differencing matrix $K$, which specifies the precision matrix of the Multivariate Gaussian sampler ($\mathbf{R} = K^T K$). Note that this modified matrix is also used in STOMP to project the noisy update \cite{Kalakrishnan:11}, however the matrix computing the smoothness term is left unchanged to enforce high smoothness at the endpoint. $\mathbf{R}$ defines a metric over trajectory space in the following way:
\begin{equation}
\begin{split}
\sum \limits_{j=1}^M \sum \limits_{i=1}^N || \ddot q_{ij} ||^2 & = (K \xi)^T (K \xi) = (\xi^T K^T) (K \xi)   \\
&  =  \xi^T (K^T K) \xi =  \xi^T \mathbf{R} \xi  = || \xi ||^2_\mathbf{R}
\end{split}
\label{eq:acc}
\end{equation}

\begin{algorithm}[t]
  {
  \label{alg:stomp}
    \SetKwInOut{Input}{input}
    \SetKwInOut{Output}{output}
		$\mathbf{R} \leftarrow K^T K$ \;
		$ \xi_t \leftarrow \mathcal{N}(\xi, \Sigma = \sigma^2 \mathbf{R}^{-1})$ \;
		\For{ i = 1 ... MaxIterations }{
			\If{ $|| x(q_{N}) - x_0 || < \epsilon $ } {
				\bf{return} $\xi_t$ \;
				}
			$C \leftarrow \left[
				\begin{array}{cccc}
					0 & ... & 0 & J(q_N)
				\end{array}
				\right]$ \;
			$\Delta \xi 
			\leftarrow - \mathbf{R}^{-1} C^T (C  \mathbf{R}^{-1}  C^T)^{-1} (x(q_{N}) - x_0)$ \;

			$ \xi_t \leftarrow \xi_t + \eta * \Delta \xi$ \;
		}
		\bf{return} Failure \;
    \caption{Goal set trajectory sampling}
    \label{GoalSetAlgo}
  } 
\end{algorithm}

\noindent
Under this metric distances between trajectories can be computed as $d( \xi_1, \xi_2) = || \xi_1 - \xi_2 ||_\mathbf{R}$. Thus to project a sample to the goal region, we must minimize the projection update with respect to that metric. This is equivalently denoted:


\begin{equation*}
\begin{aligned}
& \underset{\Delta \xi}{\operatorname{minimize}}
& \frac{1}{2} || \Delta \xi ||^2_\mathbf{R} \\
& \text{subject to}
& h(\xi_t + \Delta \xi) = 0 
\end{aligned}
\end{equation*}

\noindent
where $|| \Delta \xi ||^2_\mathbf{R} = (\xi-\xi_t)^T \mathbf{R} (\xi-\xi_t)$, and the constraint function $h$ is:
$$
h(\xi) = x(q_N) - x_0,
$$ 
\noindent
where $x_0$ is a task-space point that defines the goal set, $q_N$ is the last configuration of trajectory $\xi_t$ and $x(q)$ is the forward kinematics function for configuration $q$. The constraint function $h$ can be approximated by the first order Taylor expansion:
$$
h(\xi_t + \Delta \xi) \approx h(\xi_t) + \frac{\partial}{\partial \xi} h(\xi_t) \Delta \xi,
$$
\noindent
where $\frac{\partial}{\partial \xi} h(\xi_t) = C$ is a $P \times Q$ matrix where $P$ is the dimension of the task-space, and $Q = M * N$, is the dimension of trajectory space. $C$ contains zeros except for the last block which contains $\frac{\partial x(q_N)}{\partial q} = J(q_N) $, the kinematic Jacobian of the arm.
Thus for a small $\Delta \xi$, the Lagrangian with linearized goal constraint can be written:
$$
g(\Delta \xi, \lambda) = \frac{1}{2} \Delta \xi^T \mathbf{R} \Delta \xi + \lambda^T [ (x(q_{N}) - x_0) + C \Delta \xi ],
$$

\noindent
which solves to:

$$
\Delta \xi = - \mathbf{R}^{-1} C^T (C \mathbf{R}^{-1} C )^{-1} (x(q_{N}) - x_0),
$$

\noindent
The reader may refer to \cite{Siciliano:10} for details on this result, which is obtained by setting the gradient of the Lagrangian with respect to $\Delta \xi$ and $\lambda$ to 0.

Since the linearization of $h$ is only valid for small $\Delta \xi$, we take small incremental steps scaled by $\eta$, which is set to 0.01 in our experiments. This leads to the method presented in Algorithm \ref{GoalSetAlgo}, where $\sigma$ is the standard deviation. 

The matrix $(C  \mathbf{R}^{-1}  C^T)$ can be singular, thus in our implementation, we add a regularization term to the diagonal. Note that in our experiments, we project the samples to the joint limits while maintaining smoothness by using a QP solver. We then project the trajectory to the goal set. In order for the projection to stay within the joint limits, we modify the Jacobian matrix by zeroing out the columns that would make the update exceed the joint limits.

\begin{figure}[t]
      \center
      \includegraphics[width =0.35\linewidth]{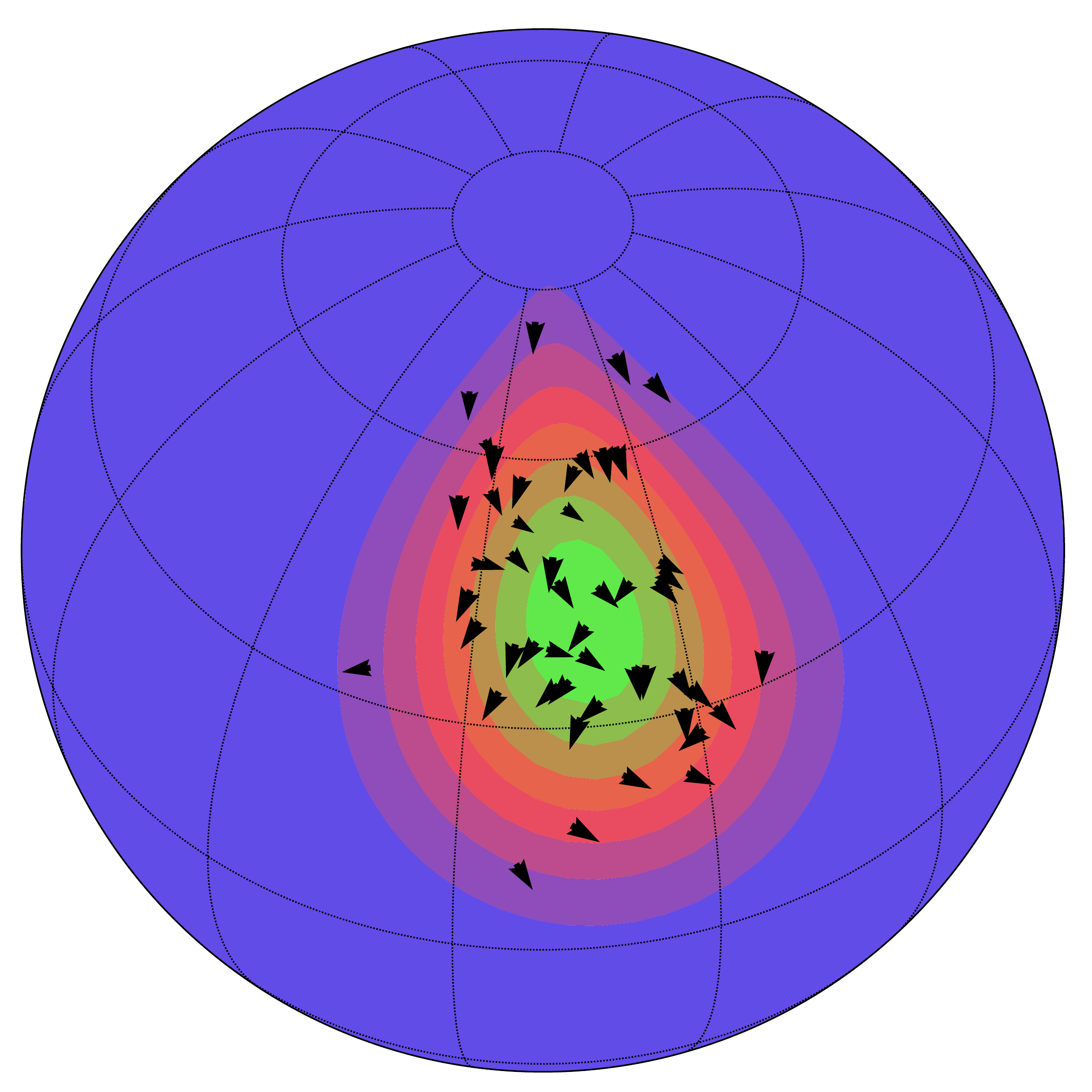}
      \includegraphics[width =0.35\linewidth]{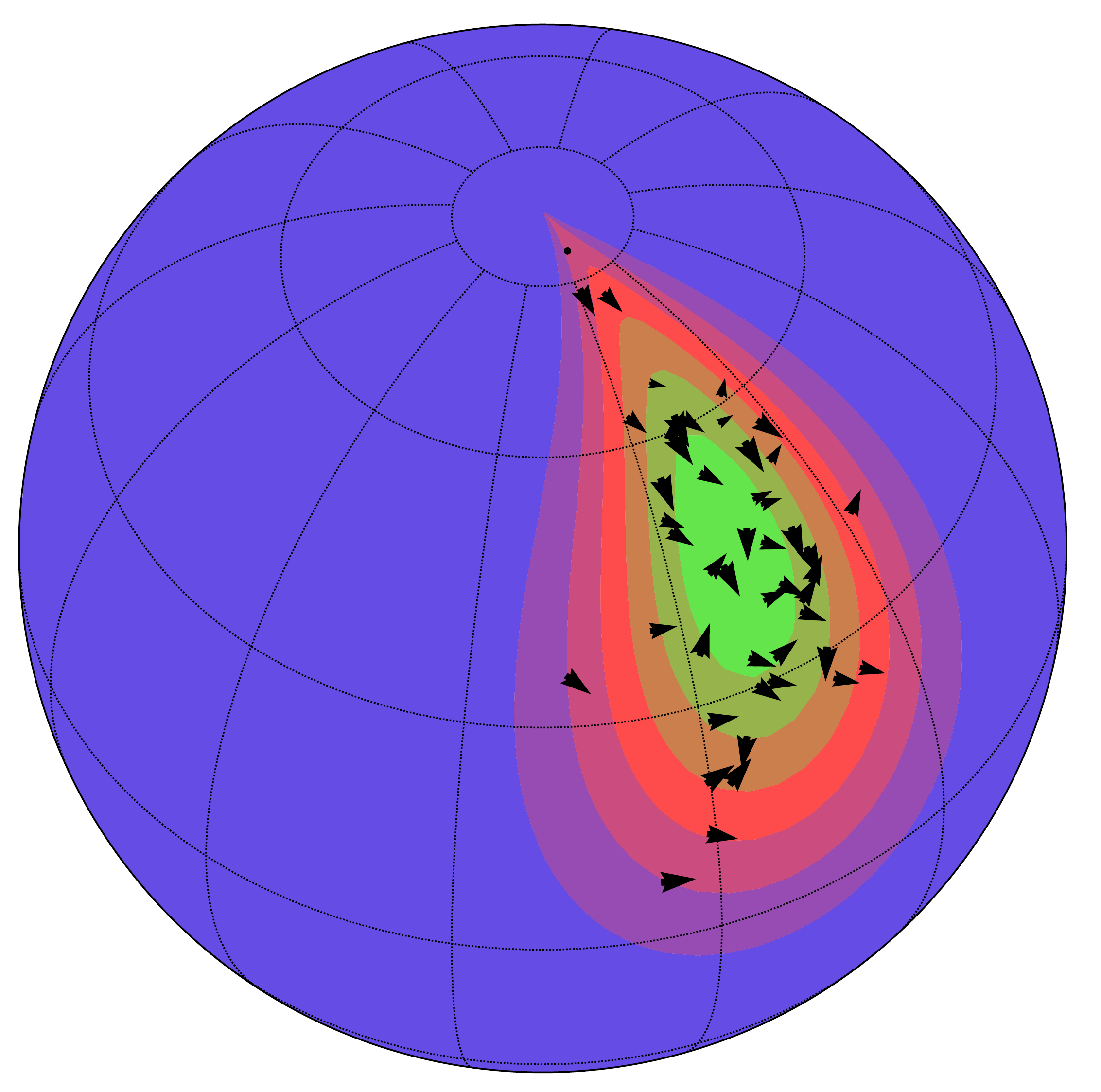}
      \caption{Goal set regions explored using the \textit{Goal set trajectory sampling algorithm} \ref{GoalSetAlgo} around two reaching motion trajectories. The color is proportional to the density of samples (green high, purple low). The standard deviation $\sigma$ is set to the value used for Inverse Optimal Control in our experiments.}
      \label{fig:goal_regions}
      \vspace{-.5cm}
\end{figure}

Figure \ref{fig:goal_regions} shows goal set regions sampled using Algorithm \ref{GoalSetAlgo}. 1000 samples are used for displaying a Gaussian kernel density estimate of the roll and pitch (azimuthal and polar angles respectively) of the frame attached to the hand at $q_N$. These two angles specify the alignment between the metacarpals long axes and the goal set center point, while the yaw angle corresponds to the orientation of the hand around that axis. A random subset of 50 samples depict yaw angles using black arrows. Note that the trajectory samples are collision free and respect joint limits.

The size of the goal set regions depends on the standard deviation $\sigma$. Larger values of the standard deviation correspond to more exploration around the goal region.

\begin{figure}[t]
      \center
      \includegraphics[width =0.50\linewidth]{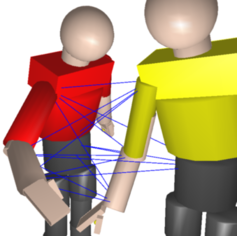}
      \includegraphics[width =0.40\linewidth]{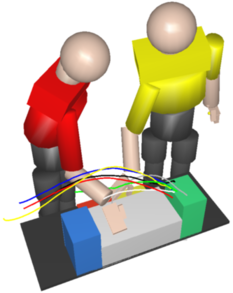}
      \caption{Each line corresponds to a distance used in the feature vector (left). 3D model of the experiment used for collision checking with hand trajectories of the seven demonstrations used in the leave-one-out test of the result section (right).  }
      \label{fig:distances}
      \vspace{-.5cm}
\end{figure}

\subsection{Human kinematic model description}
We model human kinematics following the recommendation for joints coordinates in \cite{Wu:05}. The model is composed of prismatic and hinge joints. In our experiments we only account for upper body and right arm motions, which total 23 DoF. Three translations and three rotations are used for the pelvis, three rotations for the torso joint, three translations followed by three rotations for the shoulder joint, one translation followed by three rotations for the elbow, one translation followed by three rotations for the wrist joint. 

When predicting motions, the bounds of the joints are set using the minimal and maximal values observed in the motion capture data with additional offset to allow the learning and optimization to exceed these bounds. The prismatic joints in our kinematic model are used to compensate for errors in the computation of joint centers arising from marker placement errors. They are also useful for addressing the approximations we make in modeling human kinematics.

\subsection{Feature functions}
\label{sec:features}

We consider variants of feature functions that have been introduced in previous work to account for human-robot interaction constraints \cite{Mainprice:11, Mainprice:12, Kruse:13}. We use three types of features inspired by \textit{proxemics} theory \cite{Hall:63} and experiments in neuroscience \cite{Flash:85}:

\subsubsection{Distances between human links}
The goal of these features is to avoid collision. However, in situations requiring close interaction (e.g., reaching over the other person to access an object), two people may come close to one another. To model this avoidance behavior we consider 16 pairwise distances (see Figure \ref{fig:distances}) along the arm and pelvis between the two humans (i,e,. wrist, elbow, shoulder, pelvis).

%

\subsubsection{Smoothness}
These features ensure that the trajectory remains smooth. We measure the sum of configuration and task-space length, squared velocities, squared accelerations and squared jerks along the trajectory using finite differencing.

\subsubsection{Distance to a resting posture}
These features ensure that the trajectory remains close to a resting posture by applying a weighted configuration space distance to a predefined resting posture of the form $ \sum \limits_{i=1}^N || q_i - q_{rest} ||_W \Delta t $, where $W$ is a diagonal matrix of learned weights.

\section{Human Collaboration Experiment Setup}
\label{sec:experiment}

\begin{figure}[t]

      \begin{centering}
      
      \begin{subfigure}[Initial state] {
        \includegraphics[width =0.48\linewidth]{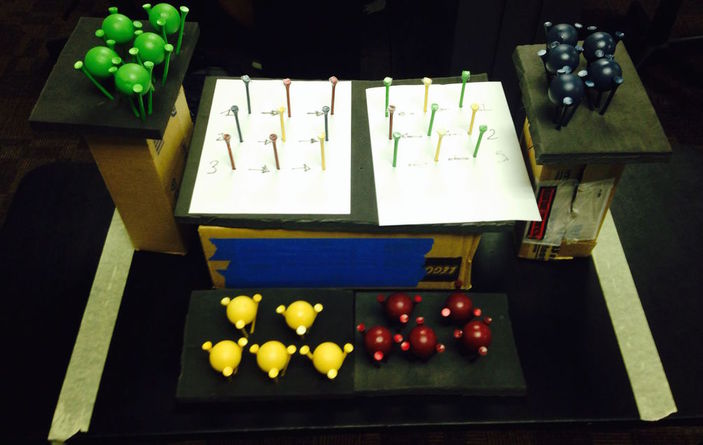}
        \label{fig:3a}
       }
       \end{subfigure}
       \vspace{-0.1cm}
       \hspace{-0.5cm}
       \begin{subfigure}[End state] {
      \includegraphics[width =0.48\linewidth]{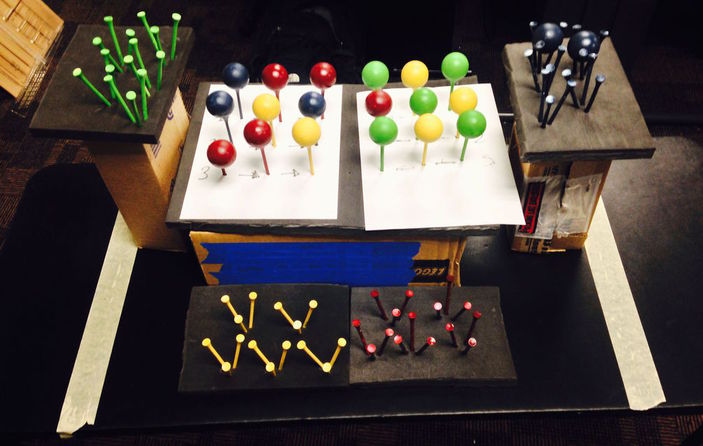}
      \label{fig:3b}
      }
      \end{subfigure}
      \vspace{-0.1cm}
      
       \begin{subfigure}[t = 0.0 sec] {
      \includegraphics[width =0.31\linewidth]{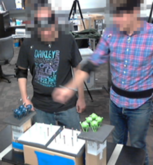}
      \label{fig:3c}
      }
      \end{subfigure}
      \hspace{-0.5cm}
      \begin{subfigure}[t = 0.5 sec] {
      \includegraphics[width =0.31\linewidth]{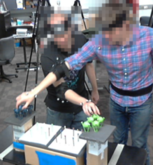}
      }
      \end{subfigure}
      \hspace{-0.5cm}
      \begin{subfigure}[t = 1.0 sec] {
      \includegraphics[width =0.31\linewidth]{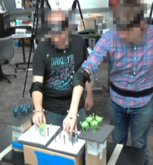}
      }
      \end{subfigure}
      
      \end{centering}
     
      \caption{Experiment design (top) and motion capture of the task (bottom). }
      \label{fig:experiment}
      \vspace{-.5cm}
\end{figure}

The aim of our experiment was to gather training and test data in a workspace sharing setting. We chose to simulate a packing task, for instance packing different chocolates into a sampler box. The experiment we designed  consisted of two participants standing shoulder to shoulder parallel to a table; each working on an individual task within a shared workspace (Figure~\ref{fig:experiment}). The task was for the participants to place colored balls on pegs of the corresponding color in a specified order (Figure~\ref{fig:3a}). 
Adhesive tape was placed on the pegs allowing quick and easy placement.

\subsection{Experiment flow}

The participants look at the color of the first empty peg in their plan, pick up a ball from the corresponding color zone, and place the ball on top of the peg, continuing until all pegs in the plan are filled with balls (Figure~\ref{fig:3b}). Following a predetermined order of execution denies the participants the ability to switch tasks in mid-motion. This allows us to study the manipulation planning component of human motion in isolation. In future work, we will investigate our results with a task planner and allow the pegs to be filled in any order.

\subsection{Recording method}

In order to record these interactions, we used a Vicon motion capture system consisting of eight Bonita cameras. 

Subjects wore a suit (seen in Figure~\ref{fig:3c}) based on standards in use in biomechanics literature \cite{Wu:05}. The suit consisted of a waist-belt and headband attached to rigid plates, a marker on the back of the hand, two on each side of the wrist, an elbow pad, two markers on either side of the shoulder, and two markers straddling both the sternum and xyphoid process. This set of markers allows us to easily find the center of rotation of the wrist, elbow, shoulder and torso. From these joint centers, we obtain a \mbox{23 DoF} configuration of the right arm and torso for each participant using analytical inverse kinematics.



\vspace{-0.1cm}
\subsection{Instructions to participants}
\label{sec:user_study}

To collect data on human interactions in a shared workspace, we conducted a human subjects study utilizing the experiment and recording methods presented in the previous sections. The study consisted of 10 pairs of participants with each pair performing the experiment 6 times for a total of 60 runs of the experiment.  The participants consisted of 4 women and 16 men with an average age of 21.  

Upon entering the experiment area, participants were read a script which briefly explained they were to perform a collaborative manipulation task. Next, the specific task to be performed was explained verbally while simultaneously being performed in front of the participants. 
Subjects were shown a resting position in which they were to hold their left arm behind their back with their right arm comfortably relaxed by their side. 
Finally, the script explained that if a part of the task was accidentally performed out of order, or if a ball fell from one of its pegs, the participant should continue performing their task instead of attempting to rectify the error. This ensured the integrity of the remainder of the task. 

\section{Human Collaboration Experiment Results}
\label{sec:results}

In this section we present results illustrating the capability of our framework to recover a cost function using the link distances, distance to a resting posture and smoothness features presented in Section \ref{sec:features}. All algorithms were implemented in C++ using the motion planning software Move3D \cite{Simeon:01}. The IOC optimization was performed in Matlab.

\paragraph{Prerequisites} In \cite{Mainprice:15}, we provide a controlled study of the approach by performing motion planning using the original STOMP algorithm on a human model with a manually defined weight vector and measure the cost difference between the initial trajectory and the recovered trajectory. This experiment gave us an estimate for the required number of trajectory samples per demonstration required by PIIRL to converge. In order to tune the regularizer, we ran IOC on the user-study dataset over a range of 10 values of the parameter and selected 0.01, which induced the best validation scores.

\paragraph{Summary}

We evaluate the quality of predictions of our goal set learning and motion algorithms against baseline tuning of the cost function by performing leave-out-testing on one class of reaching motions (i.e., reaching to a specific goal region). We then assess the method's ability to generalize across users and goal regions (three users and four goal regions) by training over a large set of motions, and we provide a comparison to Gaussian Mixture Model (GMM) based prediction on this dataset.


\paragraph{Active and Passive Humans}

We define an active human, whose motion trajectories are used as demonstrations and are later predicted, and a passive human, who may interfere with the active human. 
In the prediction phase the passive human model configuration is set from the corresponding time index of the passive human recorded trajectories. The motion planner uses the passive human configuration to generate the signed-distance-field and compute link distances between the two humans.

\subsubsection{Validation scores}

To compute the similarity between the observed trajectories and the predicted trajectories we use Dynamic Time Warping (DTW), which is an algorithm for measuring similarity between two temporal sequences that may vary in time or speed. DTW relies on a distance metric between pairs of configurations.


We use two configuration metrics throughout this section: sum of \textit{joint center distances} and \textit{task-space} distances. We do not report the configuration space metric as it does not give a fair estimate due to the high redundancy of our kinematic model, which represents the elbow and wrist joints using spherical joints.

\begin{itemize}
\item \textit{Joint center distance}: The joints considered in the first metric are the pelvis, torso, shoulder, elbow and wrist.
\item \textit{Task-space metric}: The task-space metric combines Euclidean distance and angle between consecutive Quaternions as follows:
$$
d(T_1, T_2) = \| p_1 - p_2 \| + 0.1 * cos^{-1}( | \langle v_1, v_2 \rangle | )
$$

\noindent
where $p_1$ and $p_2$ are the origins of frames $T_1$ and $T_2$ defined in some common coordinate system, and $v_1$ and $v_2$ the Quaternions.

\end{itemize}
\vspace{-0.1cm}

\begin{figure}[t]
      \center
      \includegraphics[width =1.00\linewidth]{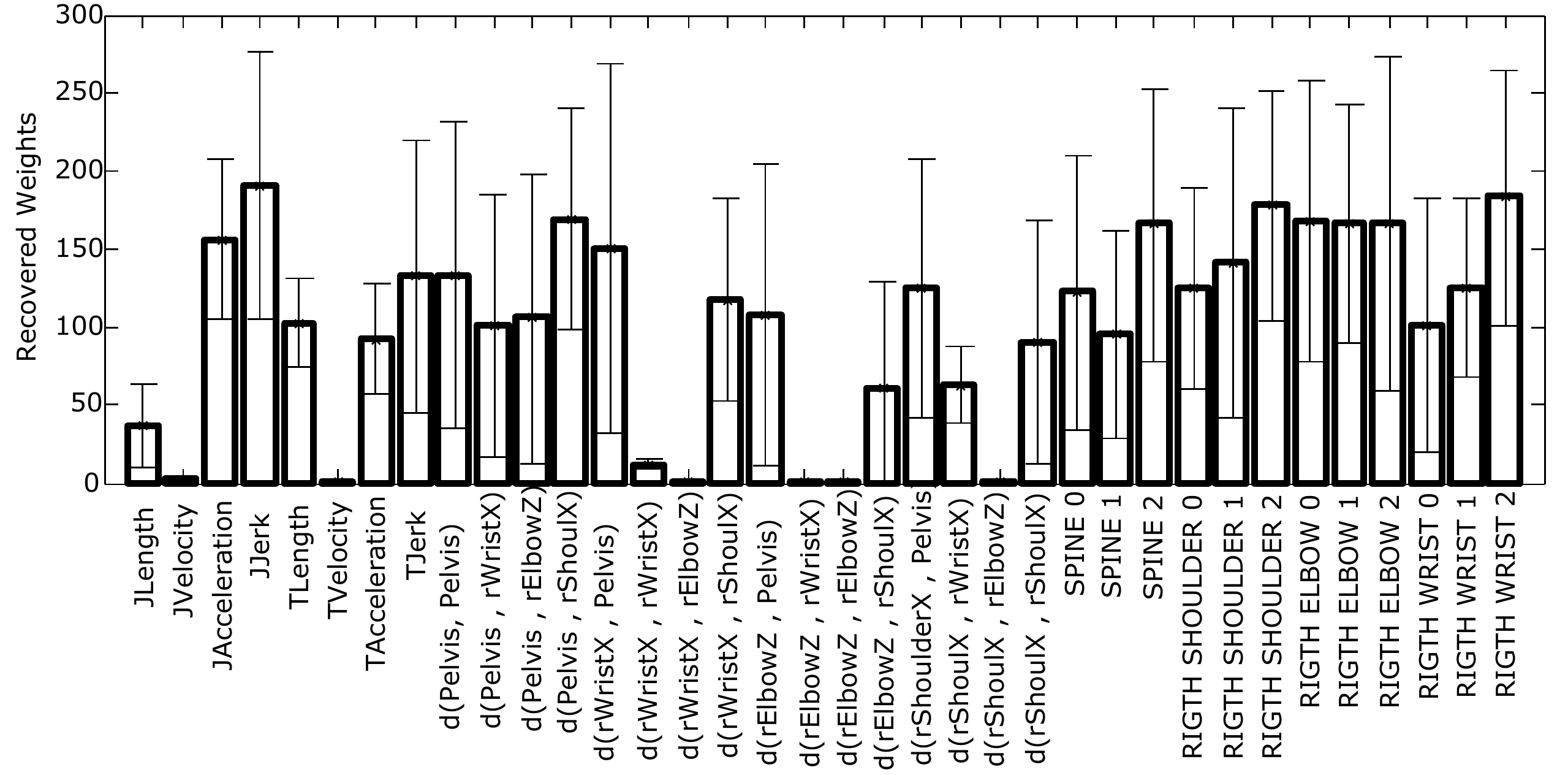}
      \vspace{-.5cm}
\caption{Mean weight vector and standard deviation from the leave-one-out testing performed over the seven motions of Figure \ref{fig:seven_motions}. The distances are described with active human in the bottom.}
      \label{fig:weights}
\vspace{-.5cm}
\end{figure}

\subsection{Evaluation}

\begin{figure*}[t]
\begin{minipage}[b]{.70\linewidth}
    \center
	\includegraphics[width =0.24\linewidth]{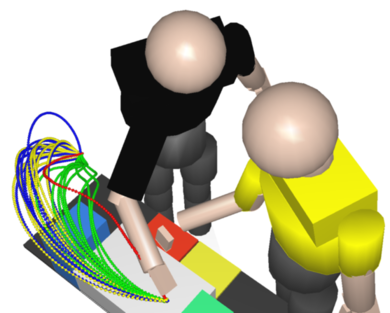}
	\includegraphics[width =0.24\linewidth]{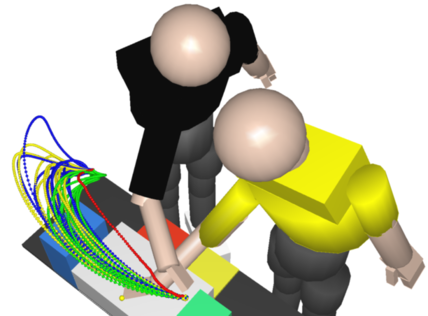}
	\includegraphics[width =0.24\linewidth]{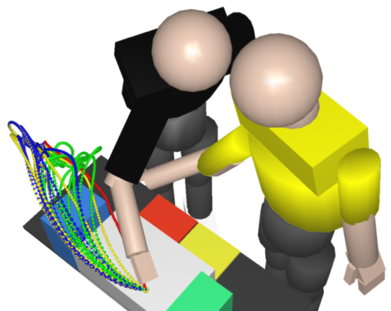}
	\includegraphics[width =0.24\linewidth]{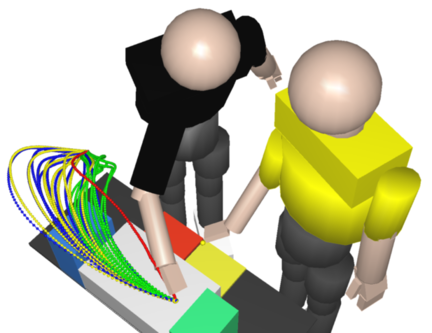}
	\includegraphics[width =0.24\linewidth]{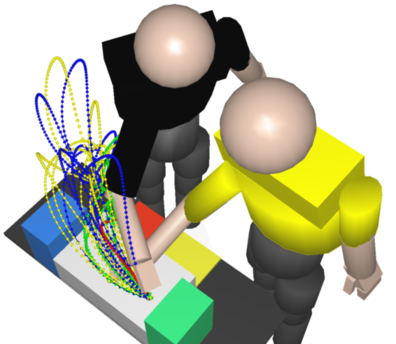}
	\includegraphics[width =0.24\linewidth]{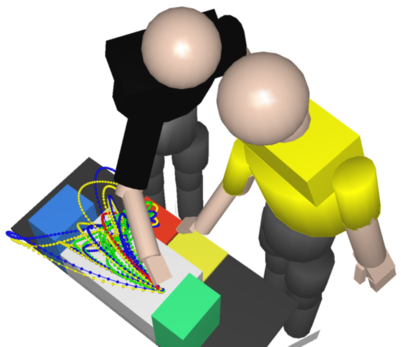}
	\includegraphics[width =0.24\linewidth]{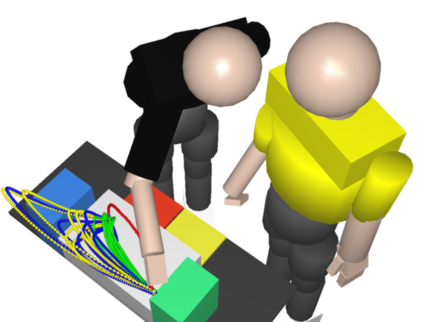}
	\caption{Seven demonstrations (red) along with the ten trajectories resulting from \textit{baseline 1} (yellow),  \textit{baseline 0} (blue) and our IOC framework (green). The demonstrations start and goal configurations are used to initialize the algorithm. The hand of the active human (black shirt) and passive human (yellow shirt) are set to the final configuration in the image.}
	\label{fig:seven_motions}
\end{minipage}
\hspace{0.05cm}
\begin{minipage}[b]{.29\linewidth}
	\center
	\includegraphics[width =0.48\linewidth]{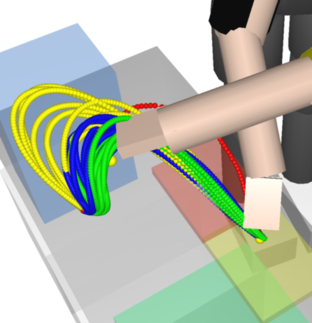}
	\includegraphics[width =0.48\linewidth]{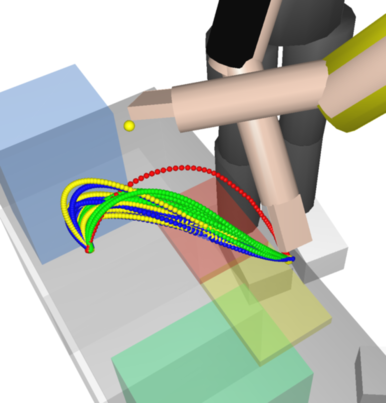}
	\includegraphics[width =0.45\linewidth]{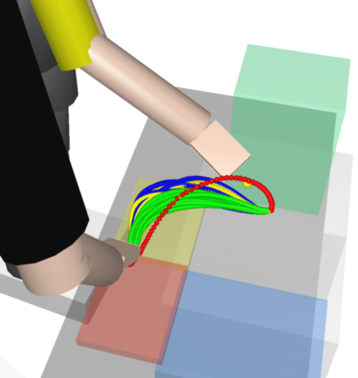}
	\includegraphics[width =0.48\linewidth]{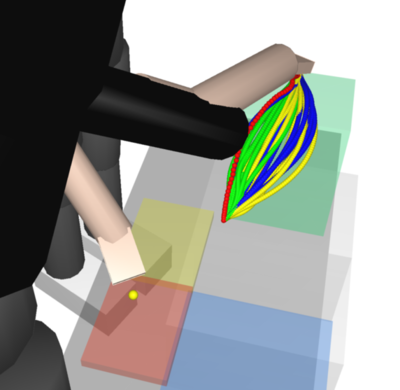}
	
    \caption{
    The IOC weight vector is learned from the dataset comprising multiple participants. Observed motion (red), \textit{baseline 1} (yellow), \textit{baseline 0} (blue) and IOC (green).
    }
    \label{fig:user_study_data}
 \end{minipage}
\end{figure*}

\begin{table*}[t]
\scriptsize
\begin{center}
\begin{tabular}{c||c|c|c|c||c|c|c|c||c|c|c|c||c|c|c|c}

\hline
& \multicolumn{8}{c|}{Re-planning} & \multicolumn{8}{c}{No Re-planning} \\
\hline
& \multicolumn{4}{c|}{task-space} & \multicolumn{4}{c|}{Joint center distances} & \multicolumn{4}{c|}{task-space} & \multicolumn{4}{c}{Joint center distances} \\
\hline
\hline

& \multicolumn{16}{c}{Dynamic Time Warping: Leave-One-Out Tests on 7 Trajectories Ending at the Same Goal Region}  \\
\hline

Method & $\mu$ & $\sigma$ & min & max & $\mu$ & $\sigma$ & min & max & $\mu$ & $\sigma$ & min & max & $\mu$ & $\sigma$ & min & max \\
\hline
\textit{baseline 1}  & 20.7 & 7.4 & 5.0 & 35.3 & 77.1 & 26.2 & 29.7 & 128.2 & 20.4 & 7.4 & 6.1 & 35.5 & 76.1 & 23.5 & 34.5 & 120.6 \\
\hline
\textit{baseline 0} & 18.3 & 6.8 & 5.1 & 37.3 & 70.1 & 22.0 & 29.7 & 118.3 & 18.3 & 6.2 & 6.0 & 31.7 & 69.7 & 20.5 & 32.6 & 111.6 \\
\hline
With IOC & 15.5 & 6.5 & 7.6 & 34.7 & 61.7 & 25.2 & 26.3 & 130.3 & 17.8 & 7.3 & 8.4 & 32.6 & 61.3 & 19.7 & 27.4 & 90.4 \\

\hline
\hline
& \multicolumn{16}{c}{Dynamic Time Warping: Generalization among Participants}  \\
\hline

Method & $\mu$ & $\sigma$ & min & max & $\mu$ & $\sigma$ & min & max & $\mu$ & $\sigma$ & min & max & $\mu$ & $\sigma$ & min & max \\
\hline
\textit{baseline 1}  &  14.4 & 14.9 & 4.0 & 64.1 &55.3 & 47.7 & 22.5 & 205.5 &  12.8 & 10.8 & 4.4 & 55.1 & 50.7 & 34.3 & 15.9 & 177.4\\
\hline
\textit{baseline 0} &  10.4 & 7.5 & 3.9 & 54.8 & 40.3 & 24.0 & 19.8 & 171.8 &  10.5 & 8.9 & 3.4 & 48.2 & 43.0 & 31.0 & 17.5 & 173.8 \\
\hline
With IOC &  9.5 & 8.9 & 3.4 & 44.9 & 46.4 & 30.2 & 20.9 & 163.4 & 9.7 & 7.9 & 3.8 & 45.3 & 46.0 & 27.3 & 19.1 & 158.9\\

\hline
\hline
& \multicolumn{16}{c|}{Spectral Analysis: Leave-One-Out Test}  \\
\hline

Method & $\mu$ & $\sigma$ & min & max & $\mu$ & $\sigma$ & min & max & $\mu$ & $\sigma$ & min & max & $\mu$ & $\sigma$ & min & max \\
\hline
\textit{baseline 1}  & 1.0 & 9.7 & -18.6 & 25.1 & 2.1 & 7.0 & -13.5 & 20.0 & -0.6 & 13.5 & -20.1 & 32.5 & 6.9 & 10.4 & -11.9 & 36.2 \\
\hline
\textit{baseline 0} & 1.2 & 11.1 & -16.6 & 33.3 & 2.2 & 7.7 & -11.4 & 24.9 & 0.9 & 15.5 & -19.9 & 49.5 & 8.4 & 12.1 & -12.2 & 47.0 \\
\hline
With IOC & 2.6 & 10.7 & -19.4 & 27.4 & -0.5 & 6.5 & -14.8 & 13.8 & 2.4 & 14.8 & -19.0 & 38.5 & 5.0 & 8.3 & -10.7 & 28.4 \\

\hline
\hline
& \multicolumn{16}{c}{Spectral Analysis: Generalization among participants}  \\
\hline

Method & $\mu$ & $\sigma$ & min & max & $\mu$ & $\sigma$ & min & max & $\mu$ & $\sigma$ & min & max & $\mu$ & $\sigma$ & min & max \\
\hline
\textit{baseline 1}  & 25.3 & 18.2 & -6.1 & 61.4 & 19.6 & 16.5 & -11.7 & 56.1 & 32.7 & 17.8 & -8.9 & 72.0 & 29.3 & 15.8 & -6.1 & 59.9 \\
\hline
\textit{baseline 0} & 30.0 & 19.0 & -3.3 & 73.7 & 23.7 & 14.2 & -9.0 & 59.9 & 34.2 & 18.1 & -0.1 & 75.8 & 32.5 & 15.7 & -6.5 & 72.2 \\
\hline
With IOC & 25.4 & 15.4 & -14.0 & 72.1 & 10.4 & 10.1 & -16.6 & 33.7 & 29.6 & 17.5 & -3.7 & 71.7 & 23.1 & 14.9 & -5.5 & 71.7 \\

\end{tabular}

\end{center}
\caption{DTW and spectral analysis performed between the observed and predicted trajectories, with and without re-planning for the leave-one-out test and the human subjects study test. Results are averaged over 10 runs.}
\label{tab:dtw_values}
\vspace{-.7cm}
\end{table*}

\subsubsection{Leave-one-out testing}

To evaluate the capability of our predictions to generalize to new situations we have performed a leave-one-out test over the seven motions of Figure \ref{fig:distances}. 
The demonstration trajectories were processed in smaller segments using the procedure described in Section \ref{sec:dyanmic}, with $\Delta t = 0.1$ sec., resulting in 33 demonstrations used for IOC. 

Each feature function is normalized to the range of the features in the samples, thus one can look at the relative influence of the parameters by looking at the weight values. The obtained mean, and standard deviation of the weights are shown in Figure \ref{fig:weights}.
This shows the relative importance of smoothness features rather than distance features, and the overall importance of the postural features. All distances between the active human's arm and passive's pelvis are important, however distances involving wrist to wrist have no influence, which is expected due to the close proximity of the two humans when sharing the workspace. The high weight values corresponding to the distances of the active's pelvis and passive's body links do not impact the overall motion as participants do not move their pelvis as much as their arm during manipulation. All the DoF are set to the bounds observed in the dataset, which constrains the pelvis motions to remain within these bounds.
Regarding the smoothness criteria, acceleration and jerk appear as dominant features, while velocity features do not appear to play any role. Unexpectedly, length of the joint space motion only plays a minor role.

For comparison, trajectories were also generated using two baseline methods:
\begin{itemize}
\item Conservative tuning (\textit{baseline 1}): the weights for the squared accelerations and 16 link distances manually set to the same value.
\item Aggressive tuning (\textit{baseline 0}): the weights for the squared accelerations set to the same value and the distance weights set to $0$.
\end{itemize}

For both baseline we do not use the postural term.
The leave-one-out section of Table \ref{tab:dtw_values} summarizes the DTW similarity values using the \textit{joint center distance} and \textit{the task-space} metric, for all methods with and without replanning. DTW is computed between the respective demonstrations (i.e., from which the initial configuration and task-space goal point are extracted) and the predicted trajectories. 

In the ``no re-planning" version Goalset-STOMP only considers the initial configuration of the passive human, while when re-planning, the passive human configuration is updated at each replanning step with the configuration at which it would be at that time step.

Figure \ref{fig:seven_motions} shows the trajectories predicted for the seven motions with each tuning method (i.e., \textit{baseline 0}, \textit{baseline 1} and IOC) but without re-planning. It also shows the trajectory executed by the human (i.e., the demonstration from which the initial configuration and task-space goal are extracted to initialize the prediction).

Trajectories planned with \textit{baseline 0} and with the IOC recovered weights have lower DTW scores than the ones planned with \textit{baseline 1}. These results are consistent throughout both metrics, with or without replanning. 
The ``no re-planning" approach tends to outperform the ``re-planning" approach slightly. This is due to the absence of motions which involved significant interference in this dataset. We report results on an example where this is not the case in Section \ref{sec:inference}.

\subsubsection{Generalization Among Participants}
\label{sec:result_user_study}

Out of the 10 pairs of participants, we selected three pairs for the quality of data obtained (i.e., absence of marker loss and occlusions of the motion capture system during the 6 runs). We manually segmented the data to obtain individual reaching motions either from a resting posture to a grasping configuration or between two grasping configurations. We selected 73 training and 20 testing motions that contained diverse start configurations and goal regions and were of good recording quality. We then learned a weight vectors using our framework by augmenting the training set as described in Section \ref{sec:approach}, leading to 461 demonstrations.

The results are reported in the lowest section of Table \ref{tab:dtw_values} and close-ups of some of the trajectories obtained without replanning, are presented in Figure \ref{fig:user_study_data}. The IOC is able to outperform the baseline methods in terms of task-space but not in terms of joint center distances in the case of \textit{baseline 0}. The standard deviations are similar to the ones found in the leave-one-out tests, but the scores are significantly lower in terms of mean. The segmented motions are in general shorter than for the leave-one-out tests, where the reaching motions go from one extreme to the other, and thus it is easier for the baseline methods to approximate human behavior using direct motions in this case.

In conclusion, for the 27 predicted motions in the three studies, in all except one test, the IOC tuned weights always outperform the baseline methods in terms of Task-space distances. Overall, it outperform the other methods in 6 out of the 8 similarity tests, where \textit{baseline 0} outperforms the others twice. This suggests that: 1) IOC with a rich set of features can be used to predict human motion in collaborative tasks and outperform simple cost functions and, 2) inter-link-distances have low impact in predicting collaborative motions. Since, the scores of using replanning do not significantly ameliorate the predictions, we could conclude that it is not necessary for these types of motion prediction in general.

\subsubsection{Significant interference}
\label{sec:inference}
However, to show the capability of the re-planning approach to better predict human motion in more difficult situations we have selected a motion where the passive human interferes significantly with the active human while he/she is reaching. 

\begin{figure}[t]
      \center
      \includegraphics[width =0.48\linewidth]{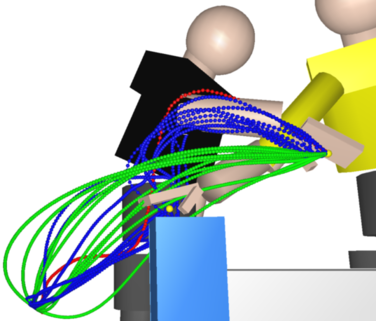}
      \includegraphics[width =0.48\linewidth]{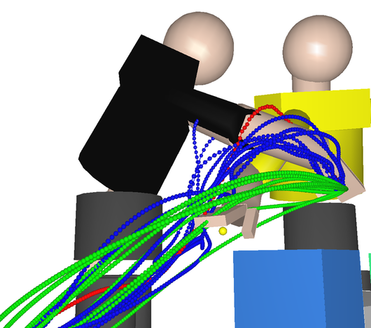}
      \caption{Two view angles of a demonstration of the benefits of re-planning on a difficult example. Original motion (red) and predicted motions with (blue) and without (green) re-planning.}
      \label{fig:trajectories}
\end{figure}

The weight vector is obtained by training with all seven motions used in the leave-one-out phase, but does not include the trajectory from which we extract the start and end configurations for prediction. The motions obtained with and without re-planning are shown in Figure \ref{fig:trajectories}, and the DTW results are shown in Table \ref{tab:re-planning}. In this case, using re-planning better predicts the active human motion because the trajectories generated with no re-planning collide with the arm of the passive human. This result is underscored by the smaller average DTW values found for the joint center distances and task-space metric. \footnote{In this example the goal set algorithms were not used, and the tuning of the task-space metric was different that used for the other experiments presented in this paper.}

\begin{table}[t]
\scriptsize
\scriptsize
\begin{center}
\begin{tabular}{c||c|c|c|c|c|c|c|c}
Type & $\mu$ & $\sigma$ & min & max & $\mu$ & $\sigma$ & min & max  \\
\hline
& \multicolumn{4}{c|}{task-space} & \multicolumn{4}{c}{Joint center distances}  \\
\hline
No re-plan & 52.8 & 9.6 & 39.9 & 67.1 & 49.2 & 8.2 & 37.7 & 63.8 \\
With re-plan & 44.9 & 6.6 & 36.1 & 55.2 &  36.2 & 8.1 & 24.81& 50.8 \\
\end{tabular}
\end{center}
\caption{DTW performed between the observed and predicted trajectories of Figure \ref{fig:trajectories}. results are averaged over 10 runs.}
\label{tab:re-planning}
\vspace{-0.8cm}
\end{table}

\subsection{Smoothness Analysis}
{\Jim
To assess the capability of our approach to produce motions which exhibit the smoothness property of human motions, we use the technique introduced in \cite{Balasubramanian:12}. This measure evaluates the spectral arc-length metric on the movement speed profile of the kinematic quantity of interest. Thus we first compute the speed profiles in task-space and joint center distances, we then compute the profiles' Fourier magnitude spectrum, which allows us to compute the spectral arc length (we use a frequency cutoff of 20 Hz and K value of 1000). In a similar fashion to what is performed for the DTW similarity measures, we report the score difference in percentage between the predicted motions and the observed motions scores for the tasks in Table \ref{tab:dtw_values}. 

Positive values indicate that the predicted motion is smoother than the human motion. 
For the leave-one-out phase, the score differences are very low indicated by the low mean and standard deviation values. This shows that the motions exhibit human-level smoothness. However for the larger dataset generated among different participants, the standard deviations are higher. In this case predicted motions tend to be much smoother than the observed human behavior (by 30\% of the smoothness measure). This can be explained by the larger diversity in the recorded motions, where the recording process was challenged by potential marker occlusions due to the interference between the participants.
}

\subsection{Comparison to GMM-GMR prediction}
{\Jim
To compare with standard motion recognition techniques using probabilistic graphical models, we implemented an algorithm based on GMM and Gaussian Mixture Regression (GMR). The 73 training examples, used in the previous study presented in Section \ref{sec:result_user_study}, were manually classified into four sets, corresponding to their goal regions. A motion for each class was computed using GMR, similarly to our prior work \cite{Mainprice:13}. 

We first train the GMM using only the active human configuration using
150 kernels for classification and 25 for regression to increase the classification rate while keeping the regressed profile smooth.
We then trained a second GMM using the passive human configuration in addition to the active human for which we used 450 kernels.

Early classifications of the 20 motions in our testing dataset were computed with 5\%, 10\%, 30\% of the trajectory execution for both cases.
We then performed DTW between the regressed motions of the class identified by the GMM classification, and the recorded trajectories. The values reported in Table \ref{fig:gmm_values}, when compared to the last rows of Table \ref{tab:dtw_values}, are significantly higher. This shows that prediction using motion planning, and IOC in particular, outperforms this approach on our data. We did not observe significant improvement when using the joint distribution (using the two humans' configurations to train and predict the motion). 
}
\begin{table}[t]


\scriptsize
\scriptsize
\begin{center}
\begin{tabular}{c||c|c|c|c|c|c|c|c}

& \multicolumn{8}{c}{GMM-GMR prediction}  \\
\hline
& \multicolumn{4}{c|}{task-space} & \multicolumn{4}{c}{Joint center distances} \\
\hline
\hline
&  \multicolumn{8}{c}{Active human used for prediction} \\
\hline

Method & $\mu$ & $\sigma$ & min & max & $\mu$ & $\sigma$ & min & max   \\
\hline
\textit{5}  & 26.2 & 14.2 & 6.3 & 54.6 & 72.2 & 38.3 & 20.0 & 158.9 \\
\hline
\textit{10}  & 25.0 & 14.2 & 6.3 & 54.6 & 68.5 & 38.4 & 20.0 & 158.9 \\
\hline
\textit{30} & 21.8 & 12.6 & 6.3 & 47.9 & 60.6 & 33.6 & 20.0 & 128.2 \\

\hline
 &  \multicolumn{8}{c}{Both humans used for prediction} \\
\hline

Method & $\mu$ & $\sigma$ & min & max & $\mu$ & $\sigma$ & min & max   \\
\hline
\textit{5} & 31.9 & 18.4 & 6.3 & 92.2 & 87.1 & 53.0 & 20.2 & 268.6 \\
\hline
\textit{10} & 20.4 & 9.9 & 6.3 & 36.8 & 54.2 & 24.9 & 20.0 & 97.7  \\
\hline
\textit{30} & 19.1 & 9.4 & 6.3 & 34.6 & 51.3 & 25.1 & 20.0 & 97.7  \\

\end{tabular}

\end{center}
\caption{DTW scores for 5, 10, and 30 \% of the trajectory}
\label{fig:gmm_values}
\vspace{-0.5cm}
\end{table}

\iftoggle{robotresults}{

\begin{figure}[t]
      \center
      \includegraphics[width =0.42\linewidth]{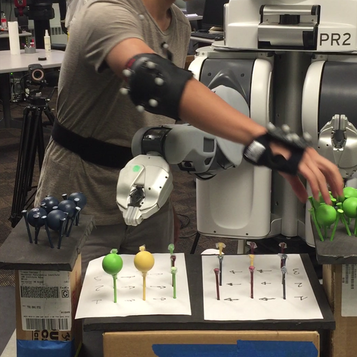}
       \includegraphics[width =0.52\linewidth]{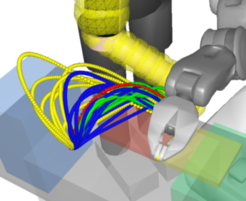}
\caption{Human-robot experiment (left). Trajectories predicted with Goalset-STOMP (right).}
      \label{fig:dtw_human_robot}
      \vspace{-0.5cm}
\end{figure}

\section{Human-Robot Workspace Sharing Experiment}
\label{sec:humanrobot}
In addition to recording human-human interactions, we created a human-robot workspace sharing experiment to evaluate the ability of this method to predict human motion when working with a robot instead of with another human. In this experiment, the human subject performs exactly the same task as described in Section~\ref{sec:user_study} while a PR2 robot executes a predetermined sequence of straight line trajectories between goal regions (see Figure \ref{fig:dtw_human_robot}). The set of robot trajectories were created with the intention of occluding the experiment workspace, while still allowing the human collaborator to complete their task.  A total of 16 subjects participated in the human-robot study of which 15 recordings produced reliable data.  In addition to being read experiment instructions, each subject performed three demo runs of their task in which the PR2 was held static to ensure proper familiarity with task execution.  Immediately following the demo runs subjects performed 8 runs of the experiment in which the PR2 executed the previously described sequence. Finally, the 120 trials of the experiment involving the PR2 were segmented into 18 trajectories forming a library of 2120 human reaching motions with a robot collaborator.

\subsection{Results}

To predict human motion with the PR2, the weight vector learned for the study presented in Section \ref{sec:result_user_study} is used and the feature function for link distances is mapped to the robot kinematics (wrist, elbow and shoulder and pelvis). 

\begin{figure}[t]
      \center
      \includegraphics[width =1.00\linewidth]{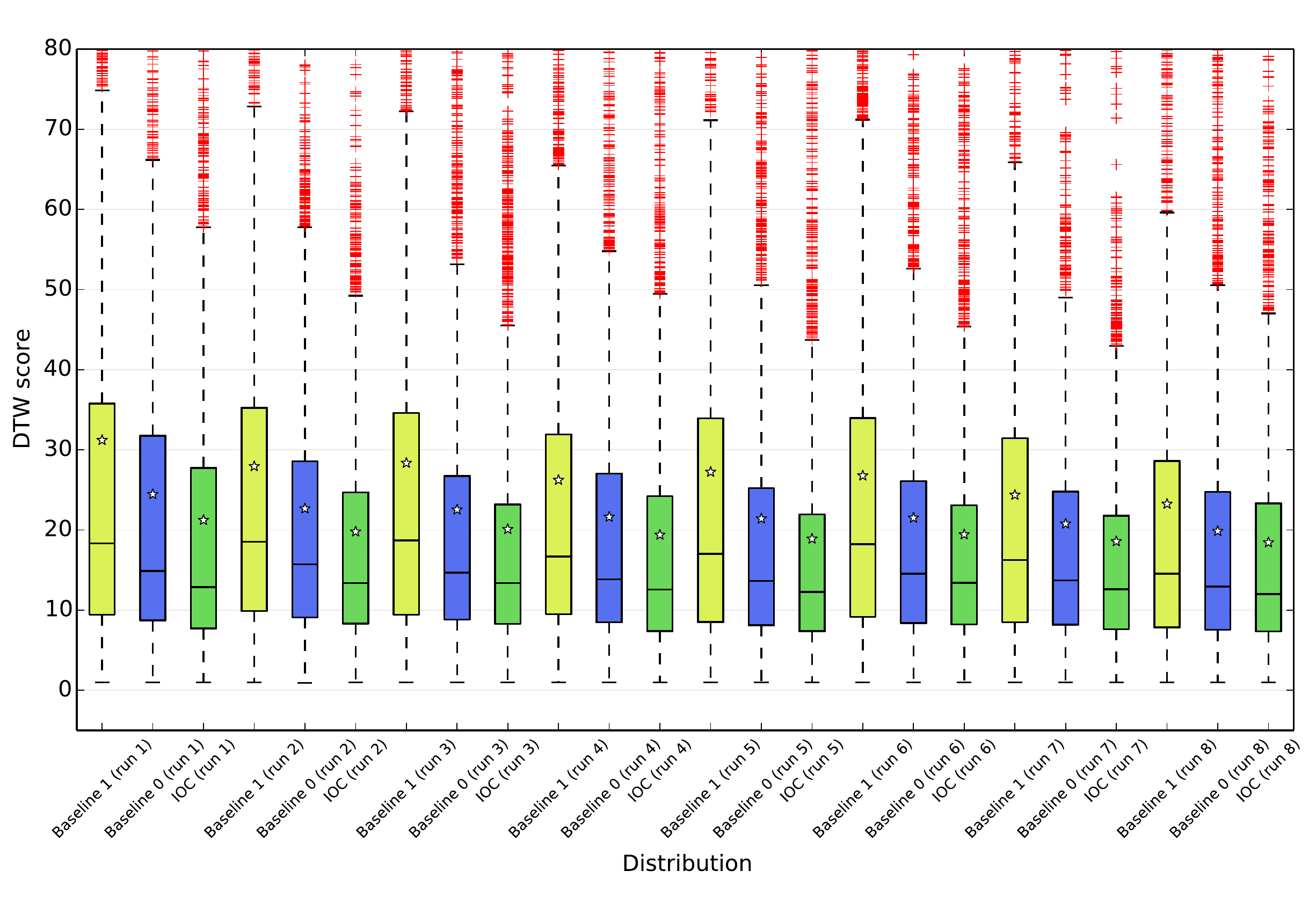}
      \vspace{-0.7cm}
      \caption{ Distribution of DTW task-space scores for the three methods of tuning, baseline 0, baseline 1, and IOC for each run.}
      \label{fig:dtw_human_robot_ioc_total_distribution}
      \vspace{-0.3cm}
\end{figure}

\begin{figure}[t]
      \center
      \includegraphics[width =1.00\linewidth]{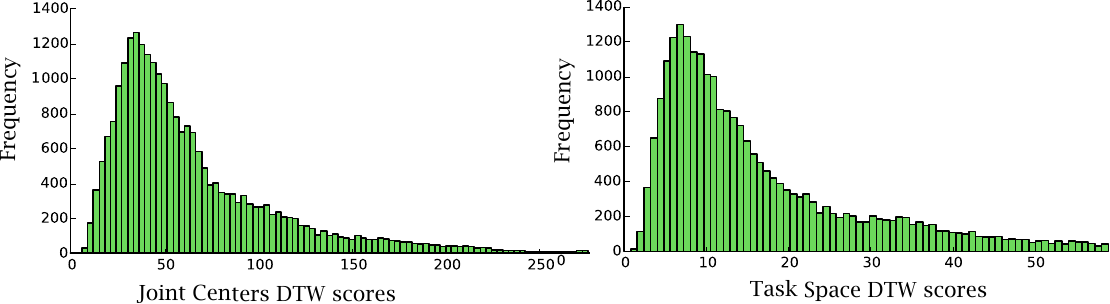}
      \vspace{-0.5cm}
	  \caption{ DTW score distributions in joint center distances and task-space observed when predicting human motions in the human-robot experiment with our framework.}
      \label{fig:dtw_human_robot_ioc_distribution}
      \vspace{-0.5cm}
\end{figure}

We ran the three tuning methods (i.e., \textit{baseline 1}, \textit{baseline 0}, and IOC) for 2120 elementary motions without replanning and we report the DTW scores distributions for each run individually in Figure \ref{fig:dtw_human_robot_ioc_total_distribution}. Note that the IOC prediction outperforms the baseline methods in each run. However many outlier exist in the distribution. These are instances where either the human hesitates or the motion planner is unable to find a collision free motion within the budgeted iterations. Figure \ref{fig:dtw_human_robot_ioc_distribution} shows the combined distribution of the IOC DTW scores for all motions. 

To examine the behavior of our prediction system as the human gets more acclimated to the robot we restrict the distribution to the first 95\% of the IOC-based method and report the mean and standard deviation for each run in Figure \ref{fig:dtw_human_robot}.
In the first run of the experiment, the participants are not acclimated to the robot and their behavior is more hesitant and thus less predictable by Goalset-STOMP, which aims to find optimal motions. Hesitations cause the human to stop, while our method assumes the human will always move to the goal. However, as the human becomes more acclimated, the predictions made by the motion planning algorithms improve, as denoted by the reduction in mean and standard deviation of the IOC-based method from the 1st (highest) to the 7th run (lowest).
}{}

\section{Conclusion and Future Work}
\label{sec:conclusion}

We have presented an important step toward predicting how humans move when collaborating on a manipulation task by applying Inverse Optimal Control to data gathered from motion capture of collaborative manipulation in a shared workspace.
To demonstrate the feasibility and efficacy of our approach we have provided test results consisting of learning a cost function, and comparing the planned motions using the learned weights to the demonstrations using Dynamic Time Warping (DTW).
The approach, based on Inverse Optimal Control (IOC) and Goal Set Iterative re-planning allows us to find a cost function balancing different features that outperforms hand-tuning of the cost function in terms of task-space and joint center distance DTW. 
\iftoggle{robotresults}{
We have also shown that our learned cost function outperforms baseline tunings of the cost function when the human works with a robot. Our prediction also improves as the human acclimates to the robot's motion.
}{}

Future work concerns enhancing the type of features to be taken into account to improve the prediction, and re-targeting these features for motion planning on a PR2 robot.

We would like to thank the Max Planck Society and Pr. Stefan Schaal for partially supporting this work, and thank Alexander Herzog and Nathan Ratliff for fruitful discussions concerning the goal set trajectory sampling algorithm.

\begin{figure}[t]
      \center
      \includegraphics[width =1.00\linewidth]{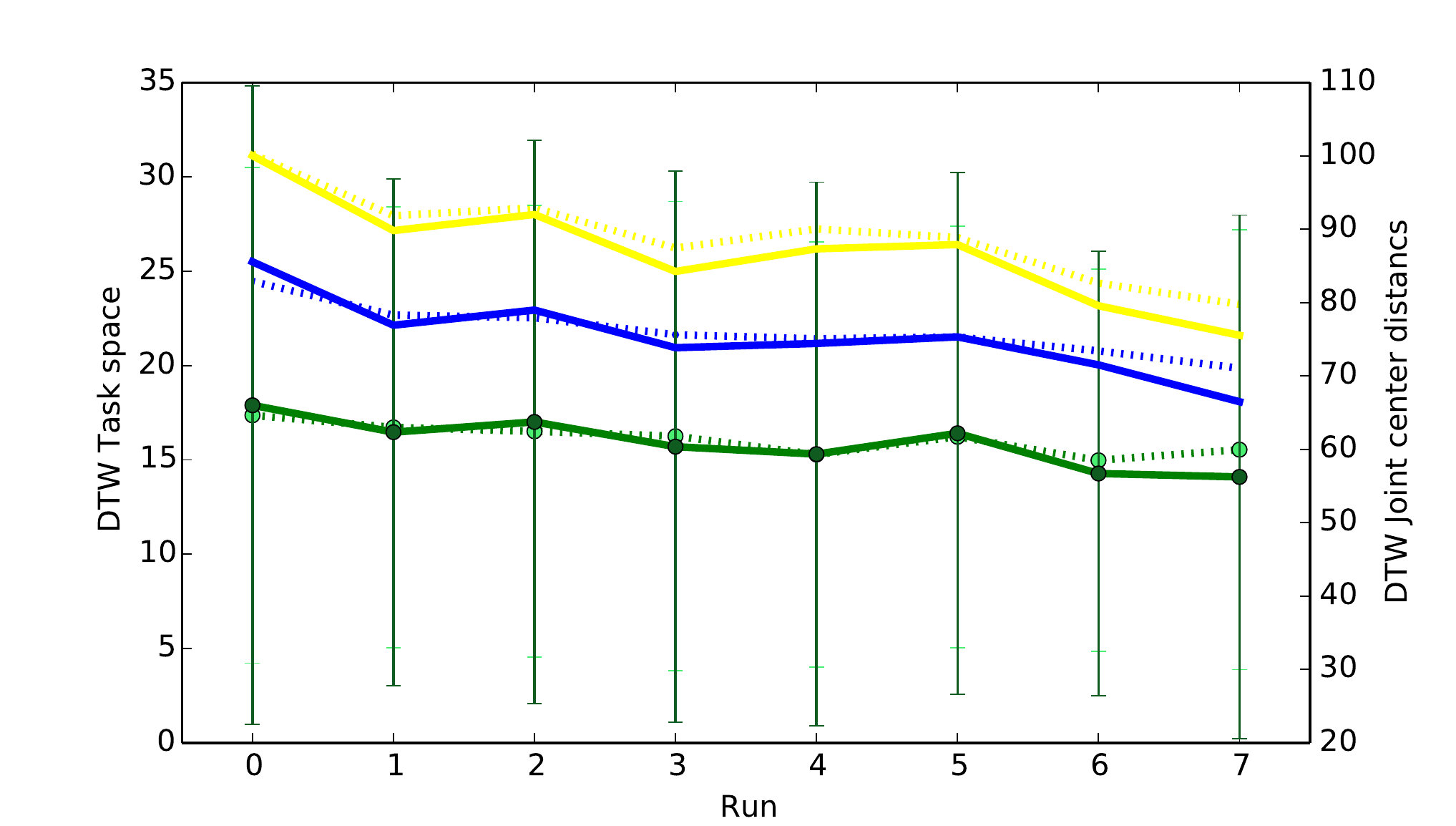}
      \vspace{-0.5cm}
\caption{Convergence of DTW scores for the IOC method (green) with the number of runs where the 5\% highest scores have been removed from the data. Baseline 0 (yellow) and baseline 1 (blue) are shown for reference. Solid lines denote joint center distance, doted lines denote task-space distance. }
      \label{fig:dtw_human_robot}
      \vspace{-0.5cm}
\end{figure}

\bibliographystyle{ieeetr}
\bibliography{root}

\vspace{-1.6cm}
\begin{biography}[{\includegraphics[width=1in,height=1.25in,clip,keepaspectratio]{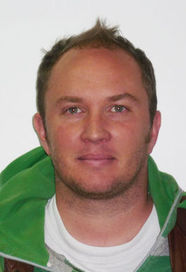}}]{Jim Mainprice} is a Postdoctoral fellow in the Autonomous Motion Department at the Max Planck Institute for Intelligent Systems (T\"{u}bingen, Germany) since January 2015. He received his M.S. from Polytech' Montpellier, France, and his Ph.D. in robotics and computer science from the University of Toulouse, France, in 2009 and 2012 respectively. His research interests include motion planning, task planning, machine learning, human-robot collaboration and human-robot interaction. 
\end{biography}

\vspace{-1.9cm}
\begin{biography}[{\includegraphics[width=1in,height=1.25in,clip,keepaspectratio]{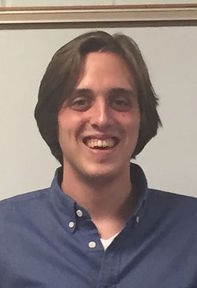}}]{Rafi Hayne}
recently received a BS in Computer Science from Worcester Polytechnic Institute where he is currently pursuing a MS degree in Computer Science.  Upon completion of this degree he hopes to continue research in robotics.  Although still developing, his research interests focus on understanding human motions as they relate to robot motion planning.

\vspace{-1.6cm}
\end{biography}
\begin{biography}[{\includegraphics[width=1in,height=1.25in,clip,keepaspectratio]{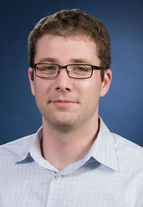}}]{Dmitry Berenson}
received a BS in Electrical Engineering from Cornell University in 2005 and received his Ph.D. degree from the Robotics Institute at Carnegie Mellon University in 2011. He completed a post-doc at UC Berkeley in 2011 and started as an Assistant Professor in Robotics Engineering and Computer Science at WPI in 2012. He founded and directs the Autonomous Robotic Collaboration (ARC) Lab at WPI, which focuses on motion planning, manipulation, and human-robot collaboration. He received the IEEE RAS Early Career award in 2016.
\end{biography}

%


\end{document}